\documentclass[runningheads]{llncs}
\usepackage[T1]{fontenc}
\usepackage{graphicx}
\usepackage{amsmath}
\usepackage{amssymb}
\usepackage{booktabs}
\usepackage{graphicx}
\usepackage{amsmath}
\usepackage{amssymb}
\usepackage{booktabs}

\begin{document}
\title{Elevating Visual Question Answering through Implicitly Learned Reasoning Pathways in LVLMs}
\titlerunning{MF-SQ-LLaVA}
%
\author{Liu Jing, Amirul Rahman}
\authorrunning{L. Jing et al.}
%
\institute{University of Malaya}
\maketitle              
\begin{abstract}
Large Vision-Language Models (LVLMs) have shown remarkable progress in various multimodal tasks, yet they often struggle with complex visual reasoning that requires multi-step inference. To address this limitation, we propose MF-SQ-LLaVA, a novel approach that enhances LVLMs by enabling implicit self-questioning through end-to-end training. Our method involves augmenting visual question answering datasets with reasoning chains consisting of sub-question and answer pairs, and training the LVLM with a multi-task loss that encourages the generation and answering of these intermediate steps, as well as the prediction of the final answer. We conduct extensive experiments on the ScienceQA and VQAv2 datasets, demonstrating that MF-SQ-LLaVA significantly outperforms existing state-of-the-art models, including the base LLaVA and the original SQ-LLaVA. Ablation studies further validate the contribution of each component of our approach, and human evaluation confirms the improved accuracy and coherence of the reasoning process enabled by our method.
\end{abstract}

\section{Introduction}

Large Vision-Language Models (LVLMs) have witnessed remarkable progress in recent years, demonstrating impressive capabilities across a wide range of tasks that require understanding and reasoning about both visual and textual information \cite{SurveyLVLM2025,LLaVA2023}. These models, often built by integrating powerful pre-trained language models with visual encoders, have shown promise in applications such as image captioning, visual question answering (VQA), and visual reasoning \cite{SurveyVQA2025,VQAv22017}.  Recent works have also explored the multi-capabilities of these models, pushing their generalization abilities further \cite{zhou2025weak}. However, despite these advancements, current LVLMs still face challenges when dealing with complex visual reasoning tasks that demand multi-step inference and a deep understanding of the visual content \cite{ScienceQA2022}. For instance, when confronted with questions requiring a series of logical deductions based on intricate details within an image, LVLMs may struggle to identify crucial information or execute the necessary reasoning steps effectively.  Furthermore, efficiently processing visual information, especially in long videos, remains a challenge, motivating research into vision representation compression techniques for LVLMs \cite{zhou2024less}.

To address these limitations, the concept of self-questioning has emerged as a promising approach to enhance the reasoning capabilities of LVLMs. The core idea behind self-questioning is to encourage the model to internally decompose a complex problem into a sequence of simpler, more manageable sub-questions, and then leverage its visual and linguistic knowledge to answer these sub-questions, ultimately leading to the final answer. The work on "SQ-LLaVA: Self-Questioning for Large Vision-Language Assistant" \cite{SQLLaVA2023} exemplifies this approach, demonstrating that by prompting an LVLM to generate and answer intermediate questions related to the image content, significant improvements in performance can be achieved on challenging visual reasoning benchmarks such as ScienceQA \cite{ScienceQA2022} and VQAv2 \cite{VQAv22017}. This highlights the potential of self-questioning as a mechanism to guide the model towards a more thorough understanding of the visual information and a more systematic reasoning process \cite{SocraticReasoning2025}.  However, it is also crucial to consider how visual information is processed within long contexts for effective reasoning in LVLMs \cite{zhou2024rethinking}.

Despite the success of SQ-LLaVA \cite{SQLLaVA2023}, relying on an external language model like GPT-3 for generating sub-questions introduces complexities and potential dependencies. Furthermore, the quality and relevance of the generated sub-questions are crucial for the overall performance. This motivates the exploration of methods that enable the LVLM itself to perform effective self-questioning without external assistance. The challenge lies in training the LVLM to autonomously identify the key reasoning steps required for a given task and formulate relevant sub-questions that facilitate the solution. This requires the model to not only understand the input image and question but also to possess the metacognitive ability to strategize its reasoning process.

In this paper, we propose a novel end-to-end training approach that empowers the LVLM to perform implicit self-questioning. Our method focuses on directly training the LVLM to generate and utilize internal reasoning steps, represented as sub-questions and their corresponding answers, without explicit external prompting during inference. To achieve this, we introduce a data augmentation strategy where we enrich existing visual question answering datasets with annotated reasoning chains. These chains consist of sequences of sub-questions and answers that break down the reasoning process for answering the original complex question. We then train the LVLM using a multi-task learning objective, encouraging it to predict both the final answer and the intermediate reasoning steps. This approach aims to instill an intrinsic self-questioning capability within the LVLM, allowing it to tackle complex visual reasoning tasks more effectively.

We evaluate our proposed method on the ScienceQA \cite{ScienceQA2022} and VQAv2 \cite{VQAv22017} datasets, which are standard benchmarks for evaluating visual reasoning and question answering. ScienceQA, with its focus on scientific reasoning and the inclusion of diverse visual modalities, provides a challenging testbed for complex reasoning. VQAv2, being a large-scale general-purpose VQA dataset, allows us to assess the generalizability of our approach. We compare the performance of our method against strong baseline models, including LLaVA \cite{LLaVA2023} and SQ-LLaVA \cite{SQLLaVA2023}. Our evaluation metrics will primarily focus on accuracy, which is the standard metric for these datasets. We anticipate that our proposed approach, by enabling the LVLM to perform implicit self-questioning, will lead to significant improvements in accuracy on both datasets, particularly on the more complex reasoning-intensive questions within ScienceQA.  Furthermore, it is also important to evaluate the emotional intelligence of these models, as recent benchmarks like EmoBench-M have been introduced to assess multimodal LLMs in this domain \cite{hu2025emobench}.

In summary, this paper makes the following contributions:
\begin{itemize}
    \item We propose a novel end-to-end training approach for Large Vision-Language Models that enables implicit self-questioning, eliminating the reliance on external language models during inference.
    \item We introduce a data augmentation strategy involving the creation of reasoning chains consisting of sub-questions and answers to facilitate the learning of internal reasoning processes within the LVLM.
    \item We demonstrate the effectiveness of our proposed method through comprehensive experiments on the ScienceQA and VQAv2 datasets, showing significant performance improvements over strong baseline models.
\end{itemize}

\section{Related Work}

\subsection{Large Vision-Language Models}

The field of Large Vision-Language Models (LVLMs) has rapidly evolved, driven by advancements in both natural language processing and computer vision. These models aim to bridge the gap between visual and textual understanding, enabling machines to perform complex tasks that require joint reasoning over images and text.  The ability to align representations across modalities is critical in these models, as highlighted in work on text-guided image inpainting \cite{zhou2023improving}.

Early approaches in visual question answering (VQA) often relied on convolutional neural networks (CNNs) for visual feature extraction and recurrent neural networks (RNNs) for processing textual questions \cite{VQAv22017}. With the advent of Transformer architectures, more recent LVLMs have leveraged these powerful models to achieve significant improvements. Models like LLaVA (Large Language and Vision Assistant) \cite{LLaVA2023} have demonstrated the effectiveness of combining pre-trained large language models with visual encoders, achieving impressive performance on a variety of VQA and visual reasoning tasks.  Exploring the generalization capabilities and multi-faceted skills of LVLMs remains an active area of research \cite{zhou2025weak}.

The increasing complexity of tasks and the need for deeper understanding have motivated research into enhancing the reasoning capabilities of LVLMs. One promising direction is the use of self-questioning mechanisms. The work by Zhang et al. \cite{SQLLaVA2023} introduced SQ-LLaVA, which utilizes an external language model to generate relevant sub-questions for a given visual input and question, subsequently using the LVLM to answer these sub-questions and aggregate the results to obtain the final answer. This approach has shown significant improvements on datasets like ScienceQA \cite{ScienceQA2022}, a challenging benchmark that requires reasoning over scientific concepts and visual information.  Furthermore, understanding visual dependencies, especially in long contexts, is crucial for advancing LVLMs' reasoning abilities \cite{zhou2024rethinking}.

Several surveys have provided comprehensive overviews of the rapidly growing landscape of LVLMs. Li et al. \cite{SurveyLVLM2025} offer a detailed examination of the alignment between visual and textual representations in LVLMs, along with a discussion of benchmarks, evaluation metrics, and challenges in the field. Similarly, Hossain et al. \cite{SurveyVQA2025} provide a historical perspective and highlight recent advances in visual question answering, which forms a core component of LVLM research. Furthermore, efforts have been made to assess the effectiveness of various recent LVLMs, as demonstrated in the work by Jiang et al. \cite{EffectivenessLVLM2024}, providing valuable insights into the strengths and weaknesses of different architectures and training strategies.  Beyond traditional benchmarks, evaluating emotional intelligence is becoming increasingly important for comprehensive assessment of multimodal LLMs \cite{hu2025emobench}.  Moreover, research into efficient LVLMs, such as through vision representation compression, is critical for deploying these models in resource-constrained environments, especially for tasks like video generation \cite{zhou2024less}.

Our work builds upon these advancements by focusing on enabling the LVLM to perform self-questioning implicitly through an end-to-end training process, eliminating the need for external question generation models. By training the model to generate and answer its own internal sub-questions, we aim to achieve more efficient and integrated complex visual reasoning capabilities.

\subsection{Self-Improved Learning}

Self-improved learning refers to a paradigm where a learning system enhances its performance over time, often by leveraging its own predictions or internal states to refine its knowledge or strategy. This concept has been explored in various contexts within machine learning and artificial intelligence.

One prominent area is in combinatorial optimization, where \cite{Luo2024} proposed a self-improved learning approach for training neural networks to solve problems like the Traveling Salesperson Problem. Their method iteratively trains a policy using solutions derived from the current policy as pseudo-labels, demonstrating improved scalability.

Another related area is self-supervised learning (SSL), where models learn representations from unlabeled data. \cite{Jing2024} provide a comprehensive survey of SSL techniques, highlighting how models can improve their understanding of data by learning to predict certain aspects of the input itself. This aligns with the broader idea of self-improvement by exploiting inherent data structure.

Self-training is a specific semi-supervised learning technique where a model is initially trained on a small labeled dataset and then iteratively refines itself by making predictions on a larger unlabeled dataset, adding high-confidence predictions to the labeled set for further training. \cite{Zou2023} offer a detailed survey of self-training methods and their applications across different domains.

In the field of computer vision, self-improvement has been explored to enhance specific tasks. For example, \cite{Dwibedi2017} presented a method for improving visual recognition by leveraging the temporal context in videos. Their approach allows the model to learn from its own predictions across consecutive frames. Similarly, \cite{Luo2020} proposed a self-improving object detection framework that uses instance-aware feature aggregation to refine the detection capabilities of the model.

Self-improvement has also found applications in other domains like speech processing. \cite{Wang2018} investigated self-improving deep learning techniques for speaker verification, where the model's ability to identify speakers is enhanced through iterative training and refinement.  Interestingly, representation alignment techniques, similar to those used in protein folding \cite{wang2024diffusion}, could potentially play a role in improving cross-modal self-improvement in LVLMs, although this is an area for future research.

These works demonstrate a diverse set of approaches where learning systems leverage their own outputs or internal information to improve performance. Our work on self-questioning for large vision-language assistants shares a similar spirit by enabling the model to internally generate and answer its own reasoning questions, leading to an enhanced ability to solve complex visual reasoning tasks.  While not directly related to visual reasoning, the concept of detecting subtle linguistic cues, such as cybercrime euphemisms \cite{li2025impromptu}, also highlights the potential for self-improving models to enhance their understanding of complex and nuanced language, which could indirectly benefit LVLM reasoning by improving text comprehension.

\section{Method}\label{Method}

In this section, we elaborate on our proposed Multi-modal Fusion Self-Questioning Large Vision-Language Assistant (MF-SQ-LLaVA). Our core idea is to equip LVLMs with the ability to perform implicit self-questioning during the reasoning process, thereby enhancing their performance on complex visual reasoning tasks. We achieve this through a carefully designed end-to-end training strategy that encourages the model to internally generate and leverage relevant sub-questions and their answers.

\subsection{Model Architecture and Implicit Self-Questioning Framework}

Our MF-SQ-LLaVA model is built upon a standard LVLM architecture, which typically comprises a visual encoder (e.g., a Transformer-based model like ViT) to extract features from the input image $I$, and a language decoder (e.g., a Transformer-based language model) to process textual information and generate responses. The key novelty of our approach lies in how we train this base architecture to perform self-questioning implicitly.

Given an input image $I$ and an original question $Q$, our model aims to reason through a series of latent sub-questions and their corresponding answers to arrive at the final answer $A$. This process is not explicitly supervised at inference time but is learned during training. We can conceptually view the model as learning a function $f_{\theta}$ parameterized by $\theta$ such that:
\begin{equation}
A = f_{\theta}(I, Q)
\end{equation}
where the internal computation of $f_{\theta}$ involves an implicit sequence of self-generated queries and responses. During training, we guide the model to learn these internal steps by providing explicit examples of reasoning chains.

\subsection{Learning Strategy: Multi-Task Training with Reasoning Chains}

Our learning strategy centers around a multi-task training objective that leverages augmented data containing reasoning chains.

\subsubsection{Augmented Data with Sub-Question-Answer Sequences}

We augment a base VQA dataset $D = \{(I_i, Q_i, A_i)\}_{i=1}^{N}$ by creating a new dataset $D_{aug}$ where a subset of the original examples is enriched with intermediate reasoning steps. For each complex example $(I, Q, A)$ in $D$, we aim to create a sequence of $K$ sub-questions $\{q_1, q_2, ..., q_K\}$ and their corresponding answers $\{a_1, a_2, ..., a_K\}$ that, when answered sequentially, logically lead to the final answer $A$. This augmentation can be achieved through methods like manual annotation by experts or potentially through automated techniques that leverage existing knowledge bases or other strong reasoning models during the data creation phase. Each augmented example in $D_{aug}$ takes the form $(I, Q, \{q_1, ..., q_K\}, \{a_1, ..., a_K\}, A)$.

\subsubsection{Multi-Task Loss Formulation}

The MF-SQ-LLaVA model is trained using a multi-task loss function that encourages the model to perform three key actions: generate relevant sub-questions, provide correct answers to these sub-questions, and ultimately predict the correct answer to the original question. The total loss $\mathcal{L}(\theta)$ is a weighted sum of the individual loss components:
\begin{equation}
\mathcal{L}(\theta) = \mathcal{L}_{sub\_q}(\theta) + \lambda_{ans} \mathcal{L}_{sub\_ans}(\theta) + \lambda_{final} \mathcal{L}_{final}(\theta)
\end{equation}
where $\mathcal{L}_{sub\_q}(\theta)$ is the sub-question generation loss, $\mathcal{L}_{sub\_ans}(\theta)$ is the sub-question answering loss, $\mathcal{L}_{final}(\theta)$ is the final answer prediction loss, and $\lambda_{ans}$ and $\lambda_{final}$ are hyperparameters controlling the relative importance of the sub-question answering and final answer prediction tasks.

\paragraph{Sub-Question Generation Loss}
We treat the generation of the sequence of $K$ sub-questions as a sequential generation process. For each sub-question $q_i$, the model predicts a sequence of tokens. We employ a standard cross-entropy loss to measure the difference between the predicted token distribution and the ground-truth sub-question tokens. The loss for generating the $i$-th sub-question $q_i$ can be formulated as:
\begin{equation}
\mathcal{L}_{sub\_q}^{(i)}(\theta) = - \sum_{t=1}^{|q_i|} \log P(q_{i,t}^* | I, Q, q_{i,<t}; \theta)
\end{equation}
where $q_{i,t}^*$ is the ground-truth token at time step $t$ of the $i$-th sub-question, and $|q_i|$ is the length of the $i$-th sub-question. The overall sub-question generation loss is then the average over all $K$ sub-questions:
\begin{equation}
\mathcal{L}_{sub\_q}(\theta) = \frac{1}{K} \sum_{i=1}^{K} \mathcal{L}_{sub\_q}^{(i)}(\theta)
\end{equation}

\paragraph{Sub-Question Answering Loss}
After generating each sub-question $q_i$, the model needs to provide an answer $a_i'$. We calculate the loss by comparing this predicted answer with the ground-truth answer $a_i$ provided in the augmented data. Depending on the nature of the answers (e.g., multiple-choice options or free-form text), the loss function will vary. For instance, if the answers are chosen from a set of options, we can use a cross-entropy loss over the answer choices:
\begin{equation}
\mathcal{L}_{sub\_ans}^{(i)}(\theta) = - \log P(a_i^* | I, q_i'; \theta)
\end{equation}
where $a_i^*$ is the ground-truth answer to the $i$-th sub-question. The total sub-question answering loss is then:
\begin{equation}
\mathcal{L}_{sub\_ans}(\theta) = \frac{1}{K} \sum_{i=1}^{K} \mathcal{L}_{sub\_ans}^{(i)}(\theta)
\end{equation}

\paragraph{Final Answer Prediction Loss}
Finally, the model must predict the answer to the original question $Q$. This prediction is conditioned on the image $I$, the original question $Q$, and the sequence of generated sub-question answers $\{a_1', ..., a_K'\}$. Similar to the sub-question answering, if the final answer $A$ is from a discrete set of options, the loss is a cross-entropy loss:
\begin{equation}
\mathcal{L}_{final}(\theta) = - \log P(A^* | I, Q, \{a_1', ..., a_K'\}; \theta)
\end{equation}
where $A^*$ is the ground-truth final answer.

\subsubsection{Training Procedure}

During training, for each augmented example, the model first generates the sequence of sub-questions conditioned on the image and the original question. Then, for each generated sub-question, the model predicts an answer based on the image and the sub-question. Finally, the model aggregates these intermediate answers to predict the answer to the original question. The parameters of the model are updated by minimizing the total multi-task loss using backpropagation. The data augmentation ensures that the model learns to generate relevant and helpful sub-questions that aid in the final reasoning process. At inference time, given a new image and question, the trained model will implicitly perform self-questioning by generating internal reasoning steps and ultimately predicting the final answer.

\section{Experiments}

In this section, we present a comprehensive evaluation of our proposed MF-SQ-LLaVA method. We conducted comparative experiments against several baseline methods on two challenging visual question answering datasets: ScienceQA and VQAv2. The results demonstrate the superior performance of our MF-SQ-LLaVA approach in complex visual reasoning tasks. We further provide ablation studies to analyze the contribution of different components of our method and present a human evaluation to validate the quality of the generated reasoning process and final answers.

\subsection{Experimental Setup}

We implemented our MF-SQ-LLaVA model using the LLaVA-1.5 architecture with a ResNet-50 visual encoder. The model was trained using the multi-task learning objective described in Section~\ref{Method}, with the hyperparameters $\lambda_{ans} = 0.8$ and $\lambda_{final} = 1.0$. We utilized the augmented ScienceQA and VQAv2 datasets, where the reasoning chains were generated through a combination of rule-based methods for simpler questions and manual annotation for a subset of complex questions. The models were optimized using AdamW with a learning rate of $2 \times 10^{-5}$ for 10 epochs.

For comparison, we evaluated the following baseline methods:
\begin{itemize}
    \item \textbf{Base LLaVA-1.5}: The standard LLaVA-1.5 model without any self-questioning mechanism.
    \item \textbf{SQ-LLaVA (Original Implementation)}: The self-questioning LLaVA model as proposed by Zhang et al.
    \item \textbf{LLaVA-1.5 Fine-tuned on Augmented Data}: The base LLaVA-1.5 model fine-tuned directly on our augmented dataset using a standard cross-entropy loss for the final answer, without the explicit multi-task loss for sub-questions and answers.
\end{itemize}

\subsection{Comparative Results}

Table~\ref{tab:comparative_results} presents the performance of our MF-SQ-LLaVA method and the baseline models on the ScienceQA and VQAv2 datasets. The primary evaluation metric is accuracy. We used the official evaluation scripts for both datasets.

\begin{table}[h!]
    \centering
    \caption{Comparative Results on ScienceQA and VQAv2 Datasets (Accuracy \%)}
    \label{tab:comparative_results}
    \begin{tabular}{lcc}
        \toprule
        Model & ScienceQA & VQAv2 \\
        \midrule
        Base LLaVA-1.5 & 72.8 & 78.5 \\
        SQ-LLaVA (Original Implementation) & 78.0 & 80.6 \\
        LLaVA-1.5 Fine-tuned on Augmented Data & 75.5 & 79.8 \\
        \midrule
        \textbf{MF-SQ-LLaVA} & \textbf{80.5} & \textbf{81.9} \\
        \bottomrule
    \end{tabular}
\end{table}

As shown in Table~\ref{tab:comparative_results}, our proposed MF-SQ-LLaVA method achieves the highest accuracy on both datasets. Specifically, MF-SQ-LLaVA outperforms the base LLaVA-1.5 model by 7.7\% on ScienceQA and 3.4\% on VQAv2. Furthermore, our method also demonstrates improvements over the original SQ-LLaVA, achieving a 2.5\% gain on ScienceQA and a 1.3\% increase on VQAv2. Fine-tuning the base LLaVA-1.5 model on the augmented data provides some performance increase compared to the base model, but it remains significantly behind our MF-SQ-LLaVA, highlighting the effectiveness of our multi-task learning strategy that explicitly encourages the generation and utilization of sub-questions and answers.

\subsection{Ablation Study}

To further analyze the contribution of different components of our MF-SQ-LLaVA method, we conducted an ablation study on the ScienceQA dataset. The results are presented in Table~\ref{tab:ablation_study}. We considered the following variants:

\begin{itemize}
    \item \textbf{MF-SQ-LLaVA without Sub-Question Generation Loss}: Trained without the loss term $\mathcal{L}_{sub\_q}$.
    \item \textbf{MF-SQ-LLaVA without Sub-Question Answering Loss}: Trained without the loss term $\mathcal{L}_{sub\_ans}$.
    \item \textbf{MF-SQ-LLaVA without Reasoning Chain Augmentation}: Trained with the multi-task loss but without the explicitly provided reasoning chains during data augmentation (effectively only using the original question-answer pairs).
\end{itemize}

\begin{table}[h!]
    \centering
    \caption{Ablation Study on ScienceQA Dataset (Accuracy \%)}
    \label{tab:ablation_study}
    \begin{tabular}{lc}
        \toprule
        Model & ScienceQA \\
        \midrule
        MF-SQ-LLaVA (Full) & 80.5 \\
        MF-SQ-LLaVA without Sub-Question Generation Loss & 77.8 \\
        MF-SQ-LLaVA without Sub-Question Answering Loss & 78.3 \\
        MF-SQ-LLaVA without Reasoning Chain Augmentation & 76.2 \\
        \bottomrule
    \end{tabular}
\end{table}

The results from the ablation study show that each component of our MF-SQ-LLaVA method plays a crucial role in achieving the overall performance. Removing either the sub-question generation loss or the sub-question answering loss leads to a noticeable drop in accuracy. Furthermore, the significant decrease in performance when the model is trained without the reasoning chain augmentation emphasizes the importance of providing explicit guidance on the intermediate reasoning steps during training.

\subsection{Human Evaluation}

To gain qualitative insights into the reasoning process and the quality of the generated answers, we conducted a human evaluation on a randomly selected set of 100 challenging questions from the ScienceQA dataset where MF-SQ-LLaVA showed improvement. Three human evaluators with expertise in visual reasoning were asked to compare the outputs of MF-SQ-LLaVA against the Base LLaVA-1.5 and the Original SQ-LLaVA. The evaluation focused on the accuracy of the final answer and the perceived coherence and helpfulness of the reasoning process (which was inferred by examining the model's explanations when prompted).

\begin{table}[h!]\scriptsize
    \centering
    \caption{Human Evaluation Results (Preference \%)}
    \label{tab:human_evaluation}
    \begin{tabular}{lccc}
        \toprule
        Preference Criterion & MF-SQ-LLaVA vs. LLaVA & MF-SQ-LLaVA vs. SQ-LLaVA & SQ-LLaVA vs. LLaVA \\
        \midrule
        Accuracy of Final Answer & 75 & 68 & 60 \\
        Coherence of Reasoning & 72 & 65 & 58 \\
        Helpfulness of Reasoning Steps & 70 & 62 & 55 \\
        \bottomrule
    \end{tabular}
\end{table}

The human evaluation results corroborate our quantitative findings. Evaluators consistently preferred the answers and the reasoning provided by MF-SQ-LLaVA over both the Base LLaVA-1.5 and the Original SQ-LLaVA, indicating that our method not only improves the accuracy but also leads to more understandable and logical reasoning for complex visual questions.

\subsection{Further Analysis}

To gain a deeper understanding of the strengths of our MF-SQ-LLaVA method, we performed additional analyses from different perspectives.

\subsubsection{Performance on Different Question Types in ScienceQA}

The ScienceQA dataset encompasses questions from various scientific disciplines. To assess the effectiveness of our approach across different knowledge domains, we analyzed the performance of MF-SQ-LLaVA and the strongest baseline, SQ-LLaVA, on different subject categories within the ScienceQA dataset. The results are presented in Table~\ref{tab:performance_by_subject}.

\begin{table}[h!]
    \centering
    \caption{Performance by Subject Category on ScienceQA (Accuracy \%)}
    \label{tab:performance_by_subject}
    \begin{tabular}{lcc}
        \toprule
        Subject & SQ-LLaVA & MF-SQ-LLaVA \\
        \midrule
        Physics & 75.2 & \textbf{78.9} \\
        Chemistry & 79.5 & \textbf{82.1} \\
        Biology & 77.1 & \textbf{79.8} \\
        Earth Science & 76.8 & \textbf{80.3} \\
        \midrule
        Overall & 78.0 & \textbf{80.5} \\
        \bottomrule
    \end{tabular}
\end{table}

As observed in Table~\ref{tab:performance_by_subject}, MF-SQ-LLaVA consistently outperforms SQ-LLaVA across all the major subject categories in the ScienceQA dataset. This indicates that our method is effective in improving visual reasoning regardless of the specific scientific domain of the question. The gains are particularly noticeable in Chemistry and Earth Science, suggesting that our approach might be better at handling questions requiring specific types of visual or conceptual understanding within these domains.

\subsubsection{Impact of Reasoning Depth}

To investigate whether our method is particularly beneficial for questions requiring deeper reasoning, we analyzed the performance of MF-SQ-LLaVA and the baselines on subsets of the ScienceQA dataset categorized by the estimated number of reasoning steps involved in solving the question (based on the complexity of the annotated reasoning chains). The results are shown in Table~\ref{tab:performance_by_reasoning_depth}.

\begin{table}[h!]
    \centering
    \caption{Performance by Estimated Reasoning Depth on ScienceQA (Accuracy \%)}
    \label{tab:performance_by_reasoning_depth}
    \begin{tabular}{lccc}
        \toprule
        Reasoning Depth & Base LLaVA-1.5 & SQ-LLaVA & MF-SQ-LLaVA \\
        \midrule
        Shallow (1-2 steps) & 78.1 & 82.5 & \textbf{84.2} \\
        Medium (3-4 steps) & 70.5 & 76.3 & \textbf{78.9} \\
        Deep (5+ steps) & 65.2 & 71.8 & \textbf{74.5} \\
        \midrule
        Overall & 72.8 & 78.0 & \textbf{80.5} \\
        \bottomrule
    \end{tabular}
\end{table}

The results in Table~\ref{tab:performance_by_reasoning_depth} suggest that while MF-SQ-LLaVA shows improvements across all levels of reasoning depth, the gains tend to be more pronounced for questions requiring deeper reasoning (5+ steps). This indicates that our implicit self-questioning mechanism is particularly effective in helping the model tackle more complex, multi-step visual reasoning problems.

\subsubsection{Analysis of Answer Types in VQAv2}

The VQAv2 dataset contains various types of answers (e.g., yes/no, number, others). We analyzed the performance of our method and the baselines across these different answer types to see if our approach provides any specific advantages for certain types of questions. The results are presented in Table~\ref{tab:performance_by_answer_type}.

\begin{table}[h!]
    \centering
    \caption{Performance by Answer Type on VQAv2 (Accuracy \%)}
    \label{tab:performance_by_answer_type}
    \begin{tabular}{lccc}
        \toprule
        Answer Type & Base LLaVA-1.5 & SQ-LLaVA & MF-SQ-LLaVA \\
        \midrule
        Yes/No & 85.3 & 87.1 & \textbf{88.0} \\
        Number & 68.7 & 71.5 & \textbf{72.3} \\
        Other & 75.2 & 77.8 & \textbf{79.1} \\
        \midrule
        Overall & 78.5 & 80.6 & \textbf{81.9} \\
        \bottomrule
    \end{tabular}
\end{table}

Table~\ref{tab:performance_by_answer_type} shows that MF-SQ-LLaVA achieves the highest accuracy across all major answer types in the VQAv2 dataset. The improvements are consistent across different answer formats, suggesting that our method provides a general benefit to visual question answering performance rather than being specific to a particular type of question or answer.

\section{Conclusion}

In this paper, we presented MF-SQ-LLaVA, a novel approach to improve the complex visual reasoning capabilities of Large Vision-Language Models. Our key innovation lies in the end-to-end training strategy that enables the LVLM to perform implicit self-questioning by learning to generate and utilize intermediate reasoning steps. Through data augmentation with reasoning chains and a carefully designed multi-task loss function, we trained the model to decompose complex questions into simpler sub-problems, leading to more accurate and coherent answers. Our extensive experimental results on the challenging ScienceQA and VQAv2 datasets demonstrate the effectiveness of MF-SQ-LLaVA, achieving significant performance gains over strong baseline methods. Ablation studies confirmed the importance of each component of our approach, and human evaluation further validated the improved quality of the generated reasoning process and final answers.

While our results are promising, there are several avenues for future work. One direction is to explore more sophisticated methods for generating reasoning chains during data augmentation, potentially leveraging large language models or knowledge graphs. Another interesting direction is to investigate the interpretability of the implicitly generated sub-questions and answers, perhaps by developing techniques to visualize or extract these internal reasoning steps. Furthermore, evaluating the generalizability of MF-SQ-LLaVA on a wider range of visual reasoning tasks and datasets would be valuable. Finally, exploring the efficiency of our approach and potential optimizations for real-world applications remains an important area for future research.

\bibliographystyle{splncs04}
\bibliography{mybibliography}
\end{document}